\documentclass{article} 
\usepackage{iclr2019_conference,times}

\usepackage[utf8]{inputenc} 
\usepackage[T1]{fontenc}    

\usepackage{hyperref}
\usepackage{url}

\usepackage{booktabs}       
\usepackage{amsfonts}       
\usepackage{nicefrac}       
\usepackage{microtype}      

\usepackage[pdftex]{graphicx}
\usepackage{epstopdf}
\usepackage{subfigure}
\usepackage{booktabs} 
\usepackage{amsmath}
\usepackage{amssymb}
\usepackage{multirow}
\usepackage{amsthm}
\usepackage{subfigure}
\usepackage{times}
\usepackage{epsfig}
\usepackage{diagbox}
\usepackage{array}
\newcolumntype{x}[1]{>{\centering\arraybackslash\hspace{0pt}}p{#1}}

\newcommand{\bfb}{\mathbf{b}}

\newcommand{\bx}{\mathbf{x}}

\newcommand{\bW}{\mathbf{W}}
\newcommand{\bw}{\mathbf{w}}
\newcommand{\bz}{\mathbf{z}}

\newcommand{\bmf}{\mathbf{f}}

\newcommand{\bbR}{\mathbb{R}}

\usepackage{xspace}

\makeatletter
\DeclareRobustCommand\onedot{\futurelet\@let@token\@onedot}
\def\@onedot{\ifx\@let@token.\else.\null\fi\xspace}

\def\wrt{w.r.t\onedot}

\makeatother

\usepackage{xcolor}

\title{Heated-Up Softmax Embedding}

\iclrfinalcopy
\author{Xu Zhang$\dagger$, Felix Xinnan Yu$\ddagger$, Svebor Karaman$\dagger$, Wei Zhang$\dagger$, \& Shih-Fu Chang$\dagger$ \\
$\dagger$ Columbia University, $\ddagger$ Google Research\\
\texttt{\{xu.zhang, svebor.karaman, wz2363, sc250\}@columbia.edu}\\
\texttt{felixyu@google.com}
}

\begin{document}

\maketitle

\begin{abstract}
Metric learning aims at learning a distance which is consistent with the semantic meaning of the samples. 
The problem is generally solved by learning an embedding for each sample such that the embeddings of samples of the same category are compact while the embeddings of samples of different categories are spread-out in the feature space. 
We study the features extracted from the second last layer of a deep neural network based classifier trained with the cross entropy loss on top of the softmax layer. We show that training classifiers with different temperature values of softmax function
leads to features with different levels of compactness.
%
Leveraging these insights, we propose a ``heating-up'' strategy to train a classifier with increasing temperatures, leading the corresponding embeddings to achieve state-of-the-art performance on a variety of metric learning benchmarks. 

\end{abstract}

\section{Introduction}

Metric learning aims at learning a metric space in which the samples from the same classes are close (compact) and the samples from different classes are far away (spread-out in the space)~\citep{hoffer2015deep, schroff_facenet, zhang2017learningb}. 
It is a fundamental research topic in machine learning and has been widely explored in a variety of computer vision applications such as 
clustering~\citep{xing2003distance,ye2007adaptive}, image retrieval~\citep{lee2008rank,zhang2017learningb}, face recognition~\citep{guillaumin2009you,schroff_facenet}, and person re-identification~\citep{koestinger2012large,lisanti2017multichannel}.

One solution for this problem is to define a loss function that enforces the properties of compactness and 
spread 
in the metric space. 
Two of the most popular loss functions are the contrastive loss~\citep{chopra_learning_2005} and the triplet loss~\citep{hoffer2015deep}. 
However, both losses face challenges in sampling, as usually there are a very large number of possible pairs or triplets in one dataset.

To overcome the sampling issue, a variety of hard mining strategies~\citep{schroff_facenet, mishchuk_working_2017, harwood_smart_2017} have been proposed.  
However, sampling the ``hardest'' samples 
(samples that most violate the predefine rules) 
often leads to poor local minima. 
Incorporating too many ``easy'' samples renders the training inefficient. 
Thus, designing a structured loss to perform hard-mining effectively and efficiently has become a hot research topic~\citep{song_deep_2016, oh_song_deep_2016}. 

Using features from the second last layer (a.k.a. bottleneck layer) of a deep neural network trained as a classifier with the softmax function and the cross-entropy loss works well for many metric learning based applications~\citep{razavian2014cnn} such as image retrieval~\citep{babenko_aggregating_2015} and face verification~\citep{liu_sphereface}. 
%
However, the goals of classifier training and metric learning are different. 
The former aims at finding the best decision function while the latter is to learn an embedding 
such that embeddings of samples of the same category are compact while those of samples of different categories are ``spread-out''.  
This motivates us to investigate the relation between metric learning and classifier training.
 
In this paper, we show that the temperature parameter in the softmax function, defined by~\citet{hinton_distilling_2015} for knowledge transfer, plays an important role in determining the distribution of the embeddings from the bottleneck layer. 
Based on the observed relation, we propose to learn a classifier with an intermediate temperature at the beginning and increase the temperature during training.
Compared to the state-of-the-art methods in deep metric learning, the proposed ``heating-up'' method achieves significantly better performance for most cases and at least comparable performances for the rest.

The main contributions of this paper are that:

\vspace{-0.7em}
\begin{itemize}
\setlength\itemsep{0.1em}
\item we study the gradient of the softmax layer and show how the temperature parameter affects the final distribution of the embedding from the bottleneck layer;
\item we propose a ``heating-up'' method that can be used to obtain an effective embedding with much better or at least comparable performance compared to state-of-the-art methods in deep metric learning.
\end{itemize}

\section{Related Works}
\label{sec:related_works}
Siamese networks with contrastive loss~\citep{chopra_learning_2005} was one of the earliest attempts to solve the metric learning problem.
By sampling either two data samples from the same category (positive pair) or two different categories (negative pair), contrastive loss tries to pull two points from positive pair together and push away the points from negative pair. 
Triplet loss~\citep{hoffer2015deep} further requires a margin between the distance of the positive pair and the distance of the negative pair. 
One of the main issues of contrastive and triplet losses is that the number of possible pairs or triplets is extremely large for a large dataset. 

A reasonable solution to address the sampling issue is mining samples that are the most informative for training, 
also known as ``hard mining''. 
There is a large body of works addressing this problem~\citep{schroff_facenet, mishchuk_working_2017, harwood_smart_2017,yuan_hard-aware_2016,wu_sampling_2017}. 
Semi-hard mining~\citep{schroff_facenet} tries to find triplets 
in a training batch, for which the distance of the positive pair and the distance of the negative pair are within a certain margin. 
HardNet~\citep{mishchuk_working_2017} is designed to mine some of the hardest triplets within one training batch. 

Designing structured losses to consider all the possible training pairs or triplets within one training batch and perform ``soft'' hard mining can be an alternative solution for hard mining~\citep{song_deep_2016, oh_song_deep_2016, ustinova_learning_2016}. 
Lifted structured loss~\citep{oh_song_deep_2016} 
exploits all triplets
in a training batch and provides a smooth loss function for hard mining. 
A few deep clustering based losses~\citep{law_deep_2017,song_deep_2016} have also been proposed to solve the problem. 
Proxy NCA~\citep{movshovitz-attias_no_2017} proposes to learn semantic proxies for training data and use a NCA loss for training. 
Applying hard mining with proxies is more efficient than with samples.

In face verification, quite a few works have shown that training a classifier and using the output of the second last layer as embedding performs reasonably well~\citep{wang_deep_2017}. 
NormFace~\citep{wang_normface} and SphereFace~\citep{liu_sphereface} suggest to $\ell_2$-normalize both the embeddings and classifier weights. 
In order to achieve promising results, a learnable or fixed scalar is usually required to be multiplied to the final logits \citep{ranjan_l2-constrained_2017}.
There are some very preliminary discussions~\citep{wang_normface, ranjan_l2-constrained_2017} about how this scalar influences the final embedding. 

This paper shows that the scalar can be seen as the temperature parameter of the softmax function 
in~\citet{hinton_distilling_2015}. 
We analyze how this temperature parameter controls the distribution, especially the compactness, of the embedding by assigning different gradients to different samples depending on their positions \wrt the boundary of the classifier. 
Inspired by these findings, we propose a ``heating-up'' strategy for training the embedding, which uses increasing temperature while training the classifier. 
The proposed method makes the embedding trained with the softmax function and the cross entropy loss achieve comparable or better performance than state-of-the-art deep metric learning methods.


\section{Revisiting Softmax Embedding with Temperature}
\label{sec:formulation}
Given a set of $n$ training samples with label $\{(\bx_1, y_1), \ldots, (\bx_n, y_n)\}$, 
where $\bx_i \in \bbR^d$ is the data sample, $d$ being the number of dimension for the training data, and $y_i \in \{1, \ldots, M\}$ is the category label of sample $\bx_i$, $M$ being the number of categories for training samples, 
we try to learn an embedding function $\bmf(\cdot): \bbR^d \rightarrow \bbR^k$, which maps a data sample to a vector in $\bbR^k$, 
such that for all $ i, j, p$ with $y_i=y_j\neq y_p$, $l(\bmf(\bx_i), \bmf(\bx_j))<l(\bmf(\bx_i), \bmf(\bx_p))$, 
where $l(\cdot, \cdot): \bbR^k \times \bbR^k \rightarrow \bbR$ is a distance function. 

We call $\bmf(\bx) \in \bbR^k$ the embedding of the data sample $\bx$ and use $\bmf$ as $\bmf(\bx)$ to simplify the notation. 
Considering training a linear classifier $\bW = [\bw_1,\ldots,\bw_M] \in \bbR^{k \times M}$ and $\bfb = [b_1, \ldots, b_M]^T \in \bbR^M$, 
$\bz = [z_1,\ldots,z_M]^T = \bW^T\bmf + \bfb \in \bbR^M$ is called the logits. 
%
The probability that sample $\bx$ belongs to category $m \in \{1, \ldots, M\}$ can be predicted by the softmax function as: 
\begin{equation}
\setlength\abovedisplayskip{5pt}
\setlength\belowdisplayskip{5pt}
p(m|\bx) = \frac{\exp(z_m/T)}{\sum_{j = 1}^{M}\exp(z_j/T)} = \frac{\exp(\alpha z_m)}{\sum_{j = 1}^{M}\exp(\alpha z_j)}
\label{eq:softmax}
\end{equation}
$T$, which is normally set to 1, is the temperature as mentioned in~\citet{hinton_distilling_2015}. 
We set $\alpha = 1/T$ as the reciprocal of the temperature to simplify the notation in the paper. 

Assuming the ground-truth distribution of the training sample is $q(m|\bx)$, generally $q(m|\bx)$ is a Dirac delta function, which equals 1 if $m=y$ and 0 otherwise, where $y$ is the ground-truth label of $\bx$, 
the cross entropy loss with respect to $\bx$, and its gradient with respect to $z_m$ are defined as: 
\begin{equation}
\setlength\abovedisplayskip{5pt}
\setlength\belowdisplayskip{5pt}
\ell(\bx,\alpha) = -\sum_{m = 1}^{M}\log(p(m|\bx,\alpha))q(m|\bx) \quad \text{and} \quad \frac{\partial \ell}{\partial z_m} = \alpha(p(m|\bx,\alpha) - q(m|\bx))
\label{eq:crossentropyloss}
\end{equation}
Considering $z_m = \bw_m^T\bmf + b_m$, we have: 
\begin{equation}
\setlength\abovedisplayskip{5pt}
\setlength\belowdisplayskip{5pt}
  \frac{\partial \ell}{\partial \bmf} =  \sum_{m = 1}^M\frac{\partial \ell}{\partial z_m} \frac{\partial z_m}{\partial \bmf} = \alpha\sum_{m = 1}^M(p(m|\bx, \alpha) - q(m|\bx))\bw_m
\label{eq:gra_zm}
\end{equation}

The proposed ``heating-up'' idea is based on an observation that different $\alpha$ values will assign 
gradients with different magnitudes to different samples and thus change the distribution of the final embeddings. 
To show this, since $\alpha$, the norm of the embedding and the norm of the classifier weights are all affecting the softmax function (Eq.(\ref{eq:softmax}) and (\ref{eq:gra_zm})), 
we follow~\citet{wang_normface, ranjan_l2-constrained_2017} and $\ell_2$-normalize the classifier weights $\bw_i$ and feature $\bmf$ to have unit norm. The normalized features and weights are denoted as $\hat{\bmf}$ and $\hat{\bw_i}$ respectively.

The following part of this section will first discuss
 how $\alpha$ changes the gradient assignment (Sec.~\ref{subsec:temperature}) and final embedding (Sec.~\ref{subsec:embedding}) with $\ell_2$-normalized embeddings and weights, then the effect of normalization (Sec.\ref{subsec:normalization}). 
Finally, we also discuss the embedding and gradient properties for an off-the-shelf classifier in~\ref{subsec:off-the-shelf}.

\subsection{Gradient Assignment by $\alpha$}
\label{subsec:temperature}

\begin{figure}[t]
\subfigure[Small $\alpha$]{\label{fig:small_alpha}
	\includegraphics[width=0.32\linewidth]{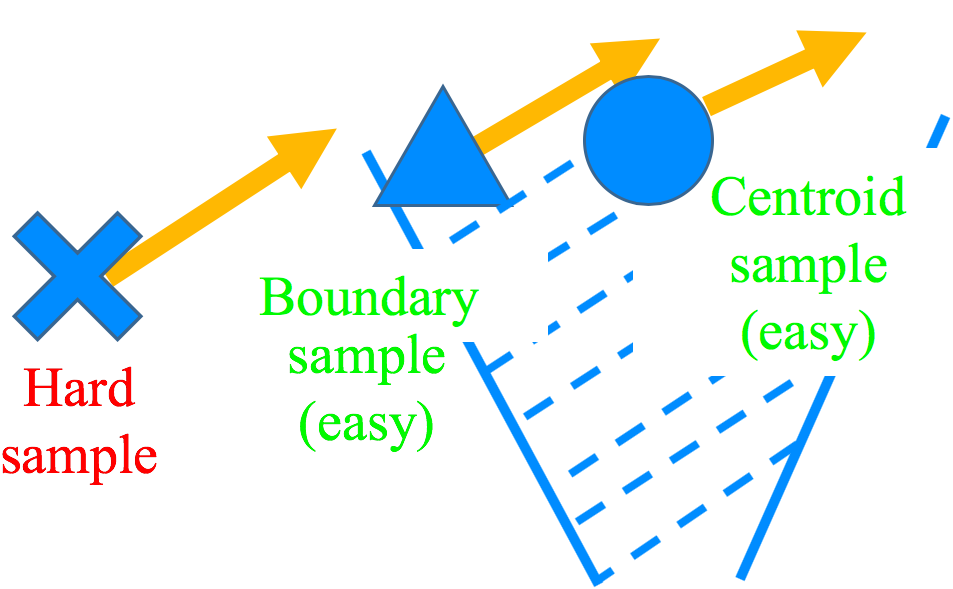}}
\hfill\vline height 70pt depth 3pt width 1pt\hfill
\subfigure[Intermediate $\alpha$]{\label{fig:medium_alpha}
	\includegraphics[width=0.32\linewidth]{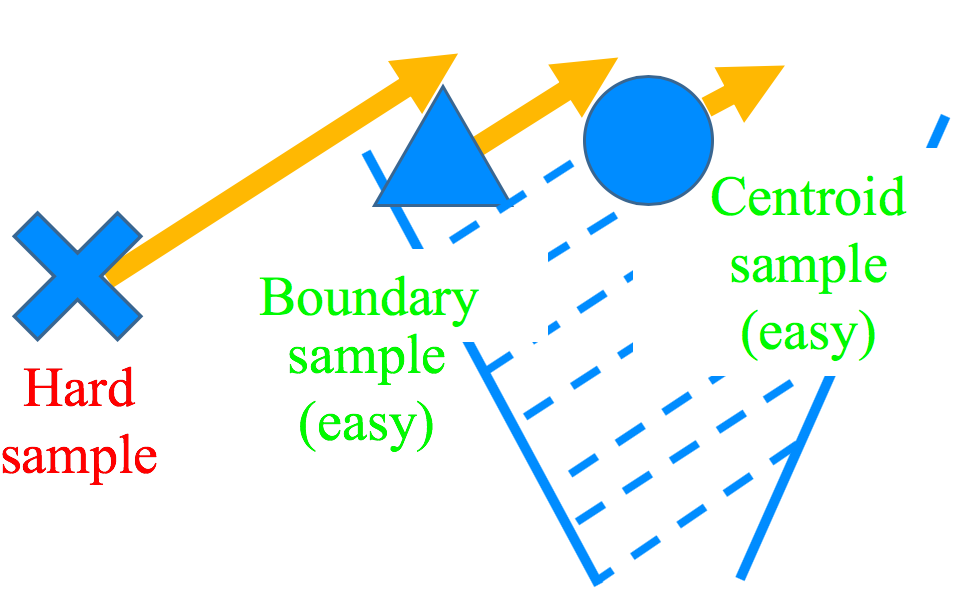}}
\hfill\vline height 70pt depth 3pt width 1pt\hfill
\subfigure[Large $\alpha$]{\label{fig:large_alpha}
	\includegraphics[width=0.32\linewidth]{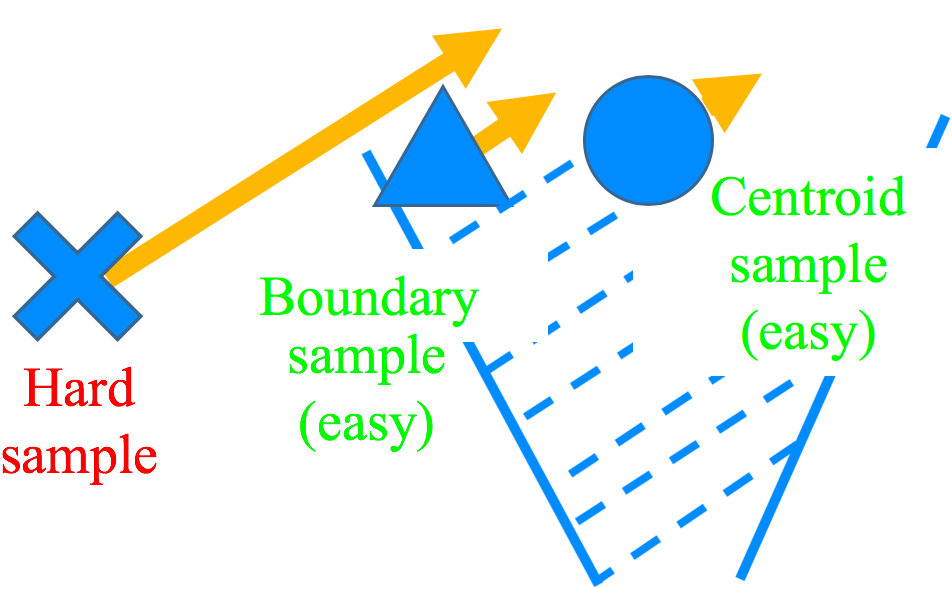}}
\vspace{-1.0em}
\caption{Different $\alpha$ values control the final embeddings by assigning different gradients to different samples. Orange arrows show gradients.}
\label{fig:alpha_overview}
\vspace{-1.0em}
\end{figure}

In this subsection, we will show how training a deep classification network with different $\alpha$ values affects the gradients of different training samples. 
From
Eq.~(\ref{eq:softmax}), when $\alpha\rightarrow+\infty$, $p(m|\bx,\alpha)$ satisfies:
\begin{equation}
\setlength\abovedisplayskip{2pt}
\setlength\belowdisplayskip{2pt}
\lim_{\alpha \rightarrow +\infty} p(m|\bx,\alpha) = 
	\left\{\begin{aligned}
	& 1/K && z_m = \max(z_1,...,z_M) \\
	& 0 && \text{otherwise, }\\
	\end{aligned}
\right.
\label{eq:limit}
\end{equation}
where $K$ is the number of logits whose value equals the maximum logits value. 
On the other hand, if $\alpha$ approaches 0, the predicted probability will approach the uniform distribution. 
In other words, as $\alpha$ increases, the predicted probability will become more ``spiky'' at the logits that have the largest value. 

We define 2 types of training samples as in Fig. \ref{fig:alpha_overview}. In the figure, all data samples (crosses, triangles and circles) belong to the same category. 
All the samples which are in the area marked by blue dashed lines will be classified as the correct category by the classifier.
The training samples can be divided into: (i) ``Hard'' samples (crosses): samples that are not classified as the correct category ($\{\bx:\exists m \neq y, z_m \geq z_y\}$); (ii) ``Easy'' samples: samples being correctly classified by the classifier ($\{\bx:\forall m \neq y, z_y>z_m\}$). 
There are two subtypes of samples in ``Easy'' category: ``Boundary'' samples~(triangles) are samples close to the decision boundary; ``Centroid'' samples 
~(circles) are samples laying close to the center of the region that belongs to the category. 

\begin{figure}[t]
\hspace{0.1\linewidth}
\subfigure[For hard samples]{\label{fig:hard_gradient}
	\includegraphics[width=0.3\linewidth]{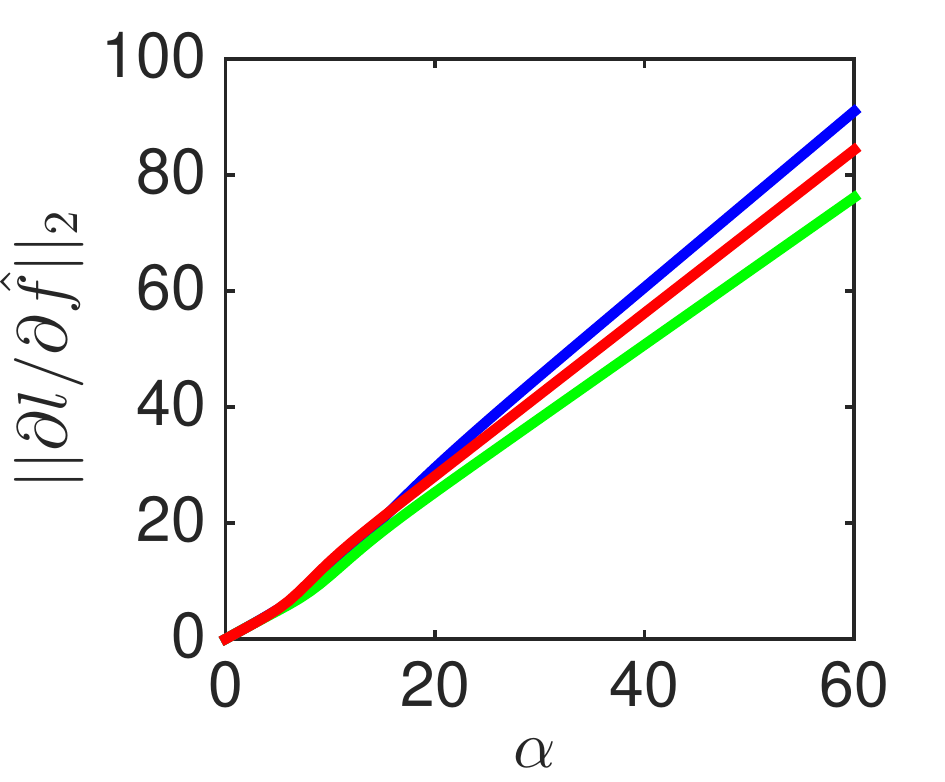}}
\hspace{0.1\linewidth}
\subfigure[For easy samples]{\label{fig:easy_gradient}
	\includegraphics[width=0.3\linewidth]{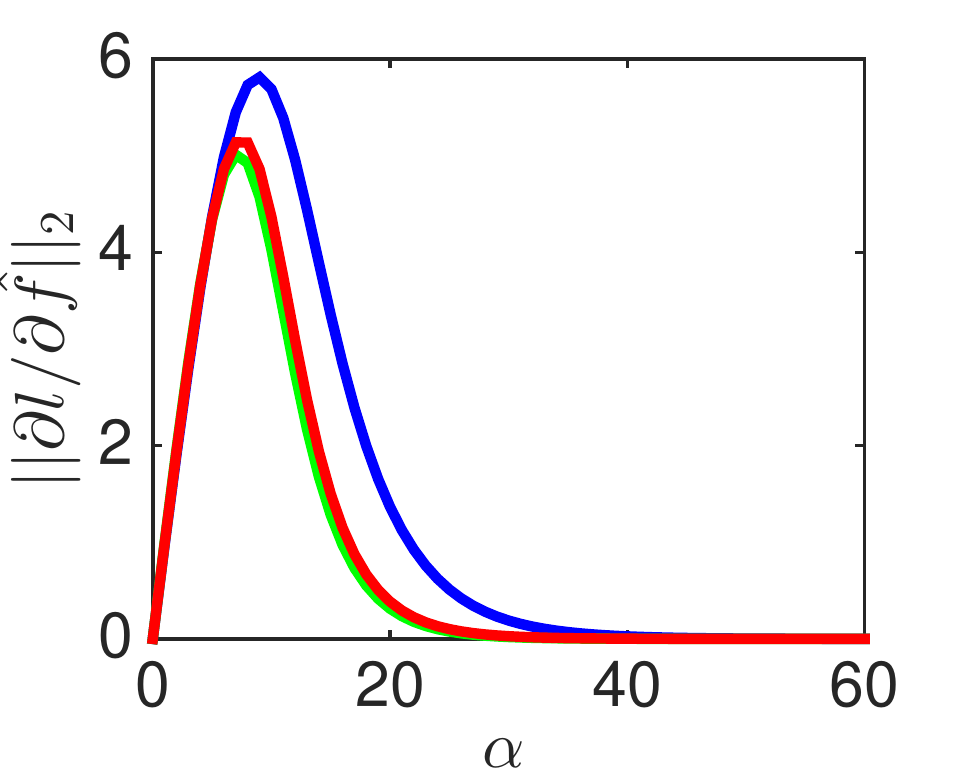}}
\hspace{0.1\linewidth}
\caption{Relation between $\alpha$ and the magnitude of the gradient with respect to embedding for both ``hard'' and ``easy'' samples.}
\label{fig:gradient}
\vspace{-1.0em}
\end{figure}

The gradient $\frac{\partial \ell}{\partial \hat{\bmf}}$ of the loss 
\wrt
the normalized embedding (i.e. Eq.~(\ref{eq:gra_zm}) with $\hat{\bmf}$ and $\hat{\bw_i}$ instead of $\bmf$ and $\bw_i$) contains $M$ terms in the sum, 
one term for each of the $M$ categories. 
There is $3$ types of terms: type 1, term with respect to the ground-truth category; type 2, term with respect to the logits that has the largest value and does not belong to the ground-truth category; type 3, other terms. We study the behavior of these terms for very small and very large $\alpha$ for ``hard'' and ``easy'' samples. 
First of all, it's easy to see that for $\alpha \rightarrow 0$, the magnitude of the gradient of any sample will approach 0.

For $\alpha \rightarrow +\infty$, considering ``hard'' samples first, the magnitude of the type 1 term ($\alpha(p(y|\bx,\alpha)-q(y|\bx))\hat{\bw}_y$) will approach $+\infty$ as $\alpha\rightarrow +\infty$, because $p(y|\bx,\alpha)$ will approach either 0 or 
$1/K$ ($K\geq2$)\footnote{$K$ can't be 1, otherwise $\bx$ is an ``easy'' sample.}
and $q(y|\bx) = 1$.
Similarly, the magnitude of the term of type 2 will also approach $+\infty$. 
For other terms, due to the property of the exponential function, that if $z_m \neq \max(z_1,...,z_M)$, $\lim_{\alpha\rightarrow + \infty}\alpha p(m|\bx,\alpha) = 0$, the magnitude of any term of type 3 will decrease to 0.
Therefore, for ``hard samples'', as $\alpha \rightarrow +\infty$, unless in some special cases\footnote{For example, two terms having exactly the same magnitude but opposite direction.}, 
the magnitude of the gradient with respect to the normalized embedding will approach  infinity. 
Fig.~\ref{fig:hard_gradient} shows the magnitude of the gradient with respect to the embedding when $\alpha$ changes for three different ``hard'' samples derived from the network in Sec.\ref{sec:experiment}.

Considering the term of type 1 for ``easy'' samples, since $\lim_{\alpha\rightarrow +\infty} p(y|\bx, \alpha) = 1$ and $\lim_{\alpha\rightarrow + \infty}\alpha (p(y|\bx,\alpha)-1) = 0$, the magnitude for this term will approach $0$. For other terms, the magnitude will also approach $0$. Therefore, the magnitude of the gradient will always approach $0$. Fig.~\ref{fig:easy_gradient} shows how the magnitude of the gradient with respect to the embedding changes with $\alpha$ for three different ``easy'' samples.

Overall, with large $\alpha$, the magnitudes of the gradients for hard samples will become very large, while the magnitudes of gradients for an easy samples will become very small (Fig. \ref{fig:large_alpha}). While, with small $\alpha$, the network will assign gradients of similar magnitudes to all the samples (Fig. \ref{fig:small_alpha}). 
Choosing an intermediate $\alpha$ value is a trade-off (Fig. \ref{fig:medium_alpha}). 
Gradient assignment for different samples will greatly affect the final embedding as discussed in the next section.

\subsection{Distribution of Final Embedding and the ``Heating-up'' Strategy}
\label{subsec:embedding}

\begin{figure}[t]
\subfigure[$\alpha=0.25$]{\label{fig:l2_norm02-embedding}
	\includegraphics[width=0.24\linewidth]{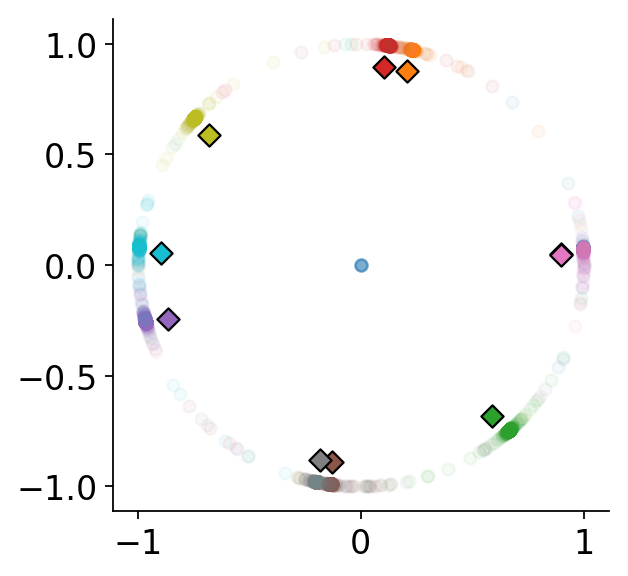}}
\subfigure[$\alpha=4$]{\label{fig:l2_norm4-embedding}
	\includegraphics[width=0.24\linewidth]{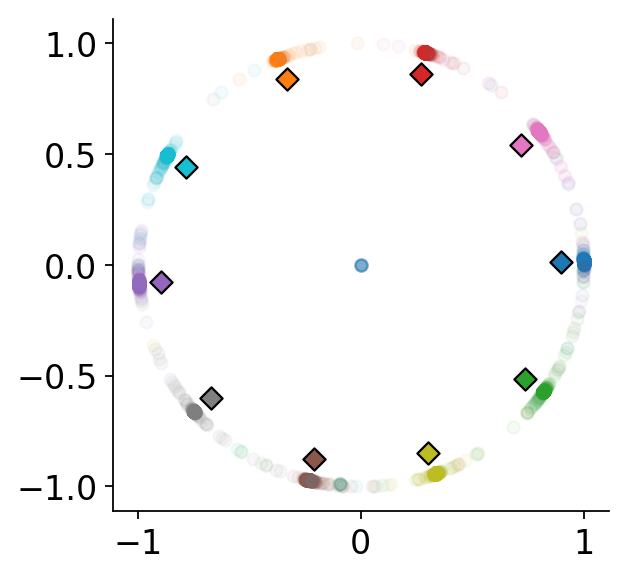}}
\subfigure[$\alpha=16$]{\label{fig:l2_norm16-embedding}
	\includegraphics[width=0.24\linewidth]{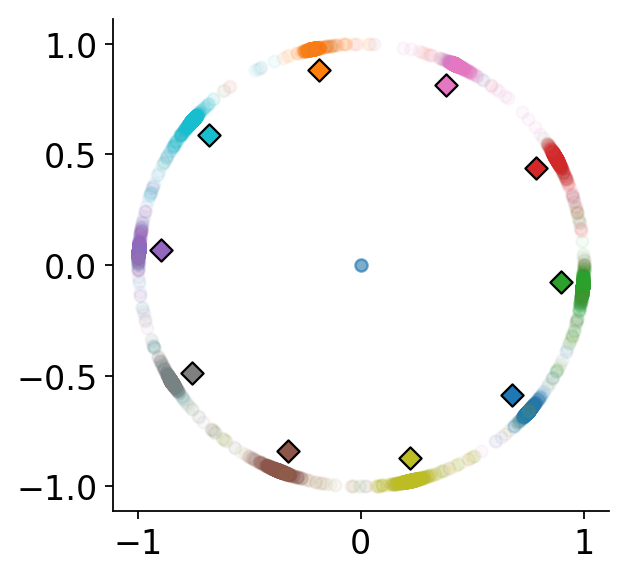}}
\subfigure[$\alpha=64$]{\label{fig:l2_norm64-embedding}
	\includegraphics[width=0.24\linewidth]{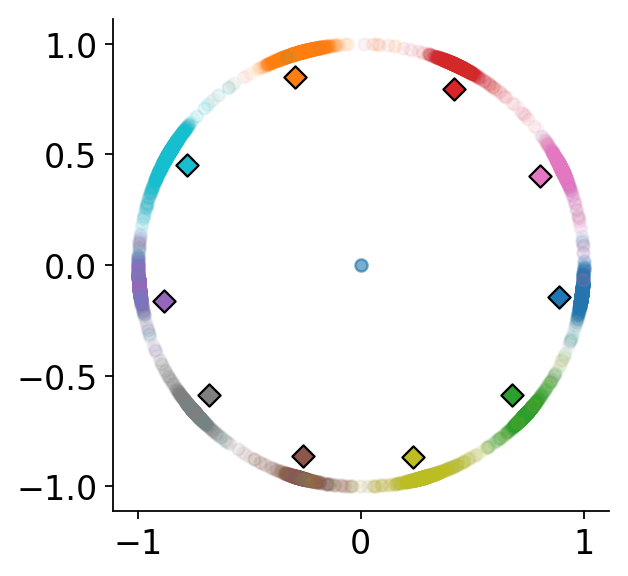}}
\vspace{-1.0em}
\caption{Embeddings trained with $\ell_2$-normalization and different $\alpha$ values.}
\label{fig:l2_embedding}
\vspace{-1.0em}
\end{figure}

To illustrate the influence of $\alpha$ on the embedding distribution, we show the embeddings obtained on the MNIST dataset in Fig.~\ref{fig:l2_norm02-embedding}-\ref{fig:l2_norm64-embedding}, when using different $\alpha$ values during training. 
Different colors represents different digits, and each diamond corresponds to the weight of the classifier of the corresponding digit. 
We slightly shift the weights towards the origin for better visualization.
The base model is LeNet~\citep{lecun2015lenet}, and the number of nodes in the second last layer is set to 2 for visualization. 
In the dataset, 50,000 samples are used for training, and 10,000 different test samples are used to draw the figure. 
All the features and classifier weights are $\ell_2$-normalized during training.

When training with small $\alpha$ (i.e. high temperature), the network will assign similar gradients to all the samples (see Fig.~\ref{fig:small_alpha}). Since the ``hard'' samples are more important to the classifier for improving the accuracy, updating ``hard'' samples and ``easy'' samples equally will make the training inefficient or even have difficulty to converge (Fig. \ref{fig:l2_norm02-embedding}).
However, choosing very large $\alpha$ (i.e. very low temperature) for training will assign large gradients to ``hard'' samples and very small gradients to all ``easy'' samples ( ``boundary'' and ``centroid''). Since the ``boundary'' samples will not get enough update, they will stay near the decision boundary. Therefore, the embedding of the samples of the same category will not be compact (Fig. \ref{fig:l2_norm64-embedding}). 
As a consequence, a good trade-off is to use an intermediate temperature for training (see Fig.~\ref{fig:medium_alpha}), where ``centroid'' samples will get small gradients, ``boundary'' samples will get intermediate gradient, and ``hard'' samples will get large gradients. 
Comparing classifiers trained with different $\alpha$ values, features of the same category are more compact for the model trained with smaller $\alpha$ values (Fig. \ref{fig:l2_norm64-embedding} $\rightarrow$ \ref{fig:l2_norm4-embedding}).

We further propose a ``heating-up'' strategy, that consists in starting training with a low or medium temperature, such that ``hard'' samples will get large gradients to update and become ``easy'' samples soon. 
After that, the temperature will be increased and therefore ``boundary'' and ``centroid'' samples will also get enough gradients for updating. 
Therefore, the final embedding of the samples of the same category will become more compact than those of the model trained with the starting temperature only. 
Multiple strategies could be defined to increase the temperature during training. 
We tried: (i) gently increasing the temperature; (ii) training with the starting temperature until convergence and using a higher temperature to fine-tune the trained network. 
Both methods lead to similar performance. 
However, since the former method would introduce an additional parameter to control the speed of increase of the temperature, we used the latter in our experiments in Sec.~\ref{sec:experiment}.

\subsection{Influence of the Normalization}
\label{subsec:normalization}
We will here discuss the influence of the normalization,  
recall that we denote the original embedding as $\bmf$ and the $\ell_2$-normalized embedding as $\hat{\bmf}$. 
The Jacobian matrix of $\hat{\bmf}$ with respect to $\bmf$ is:
\begin{equation}
\mathbf{J}_{\hat{\bmf}}(\bmf) = \frac{1}{||\bmf||_2}(\mathbf{I}-\hat{\bmf}\hat{\bmf}^T),
\label{eq:l2_norm}
\end{equation}
where $\mathbf{I}$ is the identity matrix. Considering Eq.~(\ref{eq:gra_zm}) and the chain rule, we have:
\begin{equation}
\frac{\partial \ell}{\partial \bmf} = (\frac{\partial \ell}{\partial \hat{\bmf}})^T\mathbf{J}_{\hat{\bmf}}(\bmf)
\label{eq:l2_norm_final}
\end{equation}
Considering the norm 
$||\bmf||_2$
in the denominator, the magnitude of the gradient is inversely proportional to the norm of the embedding. 
Therefore, even if the normalized embeddings are the same, the gradients \wrt the embeddings are still different for embeddings with different norms. 
The embedding with larger norm will have smaller gradient.
This may seem as a problem, and one possible solution would be to remove the norm term 
$||\bmf||_2$ 
in the denominator. 
We tried this idea for the experiments in Sec.~\ref{sec:experiment}, 
but it did not give significant improvement. 
%
The reason for that is likely that since $\partial \ell /\partial \bmf$ and $\bmf$ are always orthogonal~\citep{wang_normface}, updating the feature along the direction of gradient 
cannot change much the norm of the feature. 
We did observe that, when applying $\ell_2$-normalization to the feature during training, the norms of features before normalization are very similar. 
On the contrary, when training without 
normalization, the norm of the features may 
have large variations (see Sec.~\ref{subsec:off-the-shelf}).
For numerical stability and ease of implementation, we use Eq.~(\ref{eq:l2_norm_final}) to calculate the gradient.
Therefore, the gradient analysis in Sec.~\ref{subsec:temperature} and~\ref{subsec:embedding} still holds for unnormalized features before normalization.


We empirically found out that batch normalization~\citep{ioffe_batch_2015} without the learned scale\footnote{in this paper, batch normalization always refers to batch normalization without learned scale}, $\hat{BN}(\cdot)$, works slightly better than $\ell_2$-normalization.
We propose to define the batch normalized embedding as: 
\begin{equation}
\setlength\abovedisplayskip{5pt}
\setlength\belowdisplayskip{5pt}
\hat{\bmf}_{BN} = \hat{BN}(\bmf)/\sqrt{k}
\label{eq:bn_final}
\end{equation}
where $k$ is the number of dimensions of $\bmf$. 
Batch normalization tries to make each dimension of the embedding have zero mean and unit variance. 
Therefore, after batch normalization, the norm of the embedding is roughly $\sqrt{k}$, and the normalized feature $\hat{\bmf}_{BN}$ has norm close to 1, which is similar to $\ell_2$ normalization.
%
Batch normalization may work better than $\ell_2$-normalization because in fine-grained recognition problem, many embeddings from different categories can be very similar. 
Batch normalization removes the mean and rescales the embedding thus creating more variance. 
For classifier weights, $\ell_2$-normalization always gives us promising result.

\begin{figure}[t]
\subfigure[Embedding]{\label{fig:softmax}
	\includegraphics[width=0.234\linewidth]{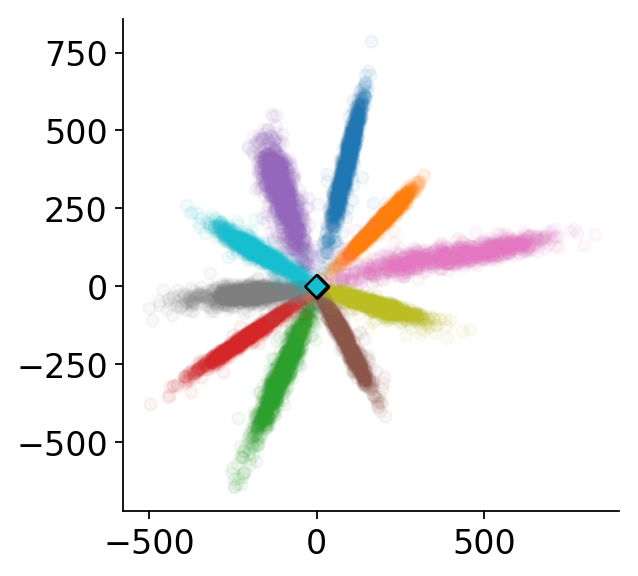}}
\subfigure[$\ell_2$-normalized embedding]{\label{fig:softmax_normed}
	\includegraphics[width=0.225\linewidth]{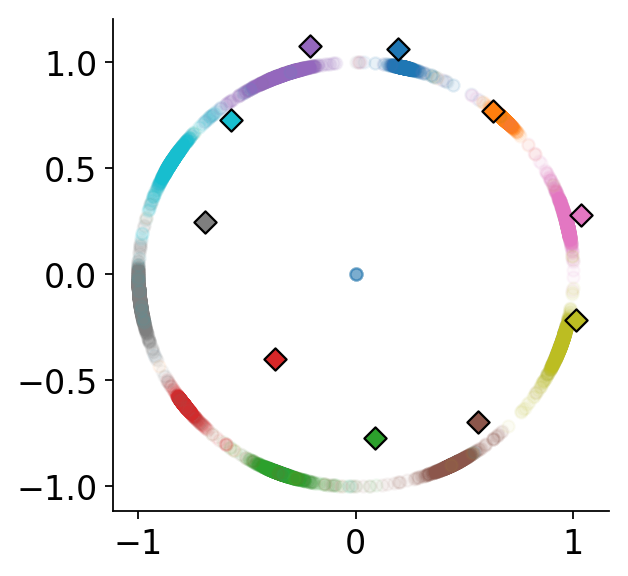}}
\subfigure[For hard samples]{\label{fig:hard_to_norm}
	\includegraphics[width=0.245\linewidth]{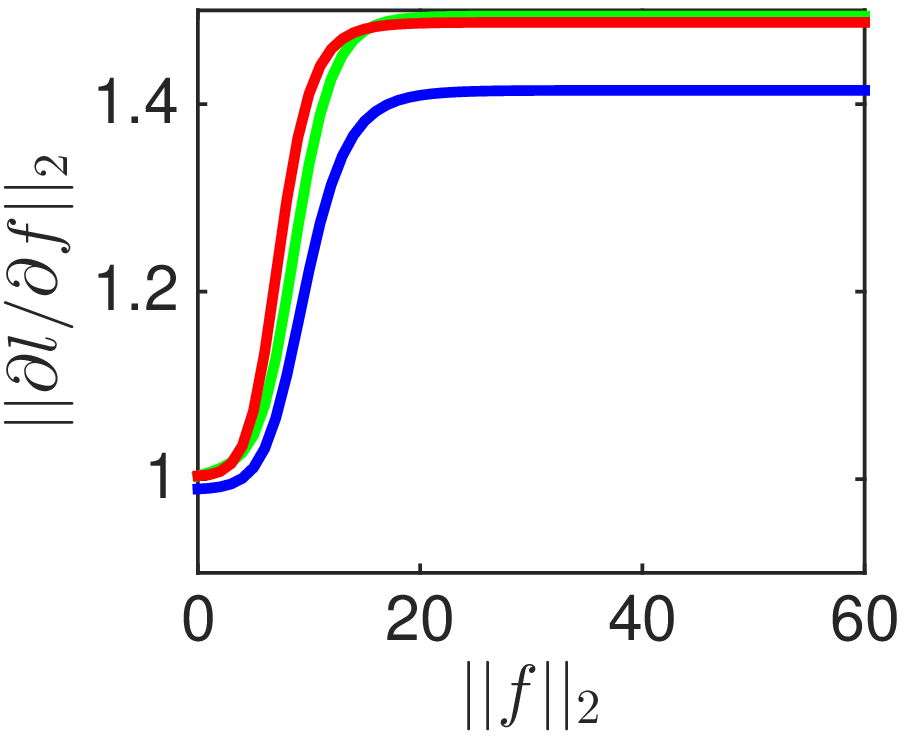}}
\subfigure[For easy samples]{\label{fig:easy_to_norm}
	\includegraphics[width=0.245\linewidth]{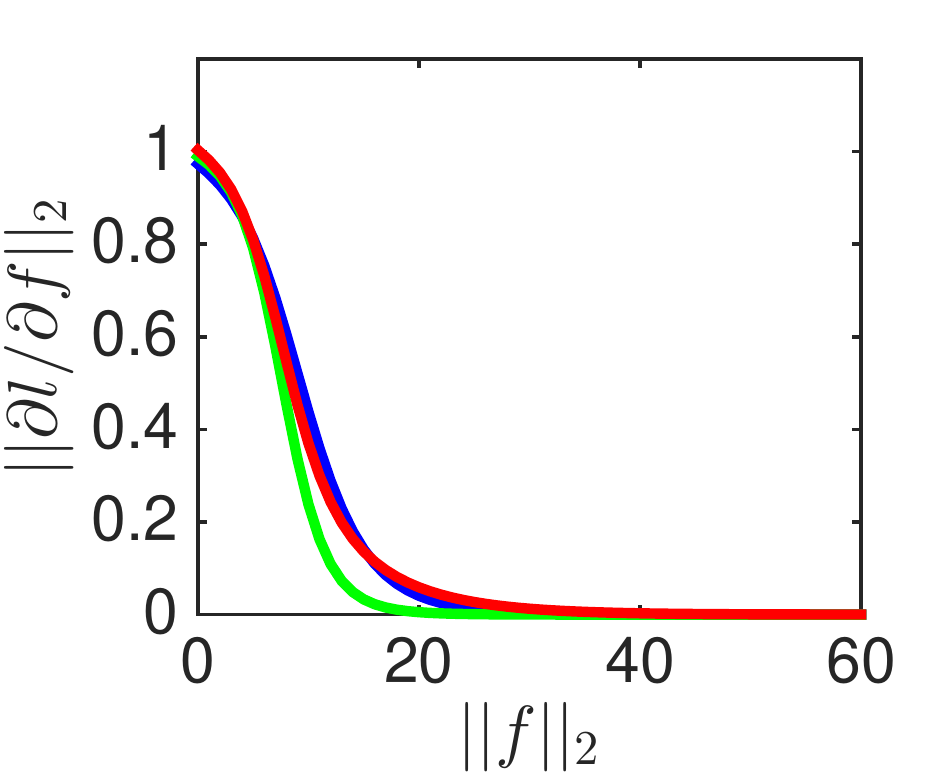}}
\caption{\ref{fig:softmax}-\ref{fig:softmax_normed}: Embedding and normalized embedding of a classifier trained without normalization. \ref{fig:hard_to_norm}-\ref{fig:easy_to_norm}: Relation between the norm of the feature and the magnitude of the gradient with respect to feature for a classifier trained without normalization.}
\label{fig:off_the_shelf}
\vspace{-1.2em}
\end{figure}

\subsection{Comparing with Off-the-Shelf Classifier}
\label{subsec:off-the-shelf}

It is interesting to compare the embedding of the proposed approach with the embedding from an off-the-shelf classifier (trained with unnormalized features and weights, $\alpha$ = 1) (see Fig.~\ref{fig:softmax} for MNIST). 
As observed in other works~\citep{wang_normface,ranjan_l2-constrained_2017}, (i) the magnitude of the embedding can be extremely large; (ii) the embedding is not ``compact''. 
Even if the feature is $\ell_2$-normalized (Fig.~\ref{fig:softmax_normed}), the embedding is still not as ``compact'' as the feature trained with normalization and proper $\alpha$~(Fig. \ref{fig:l2_norm4-embedding}). 

The norm of the embedding tends to be large because ``easy sample'' with larger norm will produce smaller loss~\citep{wang_normface}. 
To understand why the embeddings from the same category are not compact, examining the gradient is also the key. 
By setting $\alpha = 1$ in Eq.~(\ref{eq:gra_zm}), 
when $|| \bmf ||_2 \rightarrow +\infty$, for the ``easy'' sample, the magnitude of the gradient will approach~0. 
While for the ``hard'' sample, the magnitude will approach a constant, generally different from~0. 
Similarly to Fig.~\ref{fig:hard_gradient} and \ref{fig:easy_gradient}, the relation between the magnitude of the gradient \wrt the norm of the feature for some ``hard'' and ``easy'' samples is shown in Fig. \ref{fig:hard_to_norm} and \ref{fig:easy_to_norm}. 
As boundary features trained with large $\alpha$,  
boundary features with large norm will not get enough update thus leading to a not compact embedding. 

Training with $\ell_2$-normalization for the feature is different from simply applying normalization to the final feature in the test phase. 
As shown in Eq.(\ref{eq:l2_norm}) and (\ref{eq:l2_norm_final}), $\ell_2$-normalization will change the gradient of each sample during training.

\vspace{-0.7em}
\section{Experiments on Metric Learning}
\vspace{-0.5em}
\label{sec:experiment}
We conduct experiment on the following fine-grained datasets. using the training/test splits of~\citet{movshovitz-attias_no_2017}. In all these datasets, the categories in the training and test splits do \emph{not} overlap.
\vspace{-0.7em}
\begin{itemize}
\setlength\itemsep{0em}
\item \textbf{Cars} (Car196)~\citep{KrauseStarkDengFei-Fei_3DRR2013} is a fine-grained car category dataset, which contains 16,185 images of 196 car models. 8,054 images of the first 98 categories are used for training, while the 8,131 images of the other 98 categories are used for test. 
\item \textbf{Caltech-UCSD Birds-200-2011} (CUB200)~\citep{WelinderEtal2010} is a fine-grained bird category dataset which contains 11,788 images of 200 species of birds. 5,864 images of the first 100 species are used for training, while the 5,924 images of the other 100 species are used for test. 
\item \textbf{Stanford Online Product} (Product) dataset~\citep{oh_song_deep_2016} contains 120,053 images of 22,634 products categories. 59,551 images of 11,318 categories are used for training, while the other 60,502 images from 11,316 categories are kept for test.
\item \textbf{In-shop Clothes Retrieval} (Fashion) dataset~\citep{liu2016deepfashion} contains 54,642 images of 11,735 categories of fine-grained clothes. The dataset is split into 3 subsets. 52,712 images of 7,982 categories are used for training. The other 28,760 images of 3985 categories are kept for test,
split into a gallery set (12,612 images) and a query set (14,218 images). 
\end{itemize}
\vspace{-0.5em}

\subsection{Implementation Details}
We use the TensorFlow Deep Learning framework~\citep{abadi2016tensorflow} to implement the proposed method. 
For fair comparison, we exactly follow the details in~\citet{movshovitz-attias_no_2017}. 
GoogLeNet V1~\citep{szegedy2015going} 
from TensorFlow slim is used as the base network. 
The network is pre-trained with ILSVRC 2012-CLS data~\citep{ILSVRC15}. 
The input images are all resized to $256 \times 256$. For training, the resized images are randomly crop to $224 \times 224$ with random horizontal flipping. 
In test phase, we only used one single center crop as in~\citet{movshovitz-attias_no_2017}. 
For Car196, CUB200 and fashion datasets the network is fine-tuned by SGD optimizer with 0.9 momentum. 
The learning rate is set to 0.004. For product dataset, the optimizer is ADAM with learning rate of 0.01. 
The embedding size is set to 64 and the batch size is set to 32. 
We choose $\alpha = 16$ for all the datasets, as an ``intermediate'' temperature, it works well for different embedding sizes (see Sec.~\ref{subsec:embedding_size}). 
For ``heating-up'', $\alpha$ will decrease (temperature increases on the other hand) from 16 to 4 and the learning rate will decrease to 1/10 of the original learning rate. 
The training process usually converges within 50 training epochs, which is similar to the fastest state-of-the-art method, ProxyNCA~\citep{movshovitz-attias_no_2017}. Our implementation is available at: \url{https://github.com/ColumbiaDVMM/Heated_Up_Softmax_Embedding}.

\begin{table}[htb]
\caption{NMI and Recall(\%) for Car196 dataset}
\label{car_196}
\begin{center}
\begin{small}
\begin{sc}
\begin{tabular}{p{0.73cm}|p{0.73cm}p{0.73cm}p{0.73cm}p{0.73cm}p{0.73cm}p{0.73cm}|p{0.73cm}p{0.73cm}p{0.73cm}p{0.73cm}p{0.73cm}}
\toprule
 & [1] & [2] & [3] & [4] & [5] & [6] & SM & LN & BN & HLN & HBN \\
\midrule
NMI    & 53.35 & 56.88 & 54.44 & 61.12 & 59.50 & 64.90 & 59.52 & 62.40 & 65.81 & 66.87 & \textbf{68.10}\\
R@1    & 51.54 & 52.98 & 58.11 & 67.54 & 64.65 & 73.22 & 60.76 & 68.59 & 71.12 & 71.93 & \textbf{74.70}\\
R@2    & 63.78 & 66.70 & 70.64 & 77.77 & 76.20 & 82.42 & 73.58 & 78.55 & 80.62 & 81.68 & \textbf{83.90}\\
R@4    & 73.52 & 76.01 & 80.27 & 85.74 & 84.23 & 86.36 & 82.50 & 86.18 & 87.82 & 88.34 & \textbf{89.77}\\
\bottomrule
\end{tabular}
\end{sc}
\end{small}
\end{center}
\vskip -0.1in
\end{table}

\begin{table}[htb]
\caption{NMI and Recall(\%) for CUB200 Dataset}
\label{cub200}
\begin{center}
\begin{small}
\begin{sc}
\begin{tabular}{p{0.73cm}|p{0.73cm}p{0.73cm}p{0.73cm}p{0.73cm}p{0.73cm}p{0.73cm}|p{0.73cm}p{0.73cm}p{0.73cm}p{0.73cm}p{0.73cm}}
\toprule
 & [1] & [2] & [3] & [4] & [5] & [6] & SM & LN & BN & HLN & HBN \\
\midrule
NMI    & 55.38 & 56.50 & 59.23 & 56.87 & 59.90 & 59.53 & 57.19 & 59.23 & 59.20 & 60.34 & \textbf{60.75}\\
R@1    & 42.59 & 43.57 & 48.18 & 50.08 & 49.78 & 49.21 & 44.02 & 46.86 & 47.27 & 49.68 & \textbf{50.68}\\
R@2    & 55.03 & 56.55 & 61.44 & 62.24 & 62.34 & 61.90 & 55.86 & 59.79 & 59.67 & 61.85 & \textbf{62.58}\\
R@4    & 66.44 & 68.59 & 71.83 & 73.38 & \textbf{74.05} & 67.90 & 68.18 & 71.56 & 71.89 & 73.08 & 73.82\\
\bottomrule
\end{tabular}
\end{sc}
\end{small}
\end{center}
\vskip -0.1in
\end{table}

\begin{table}[htb]
\caption{NMI and Recall(\%) for Stanford Product Dataset}
\label{product}
\vskip 0.05in
\begin{center}
\begin{small}
\begin{sc}
\begin{tabular}{l|ccccc|ccccc}
\toprule
 & [1] & [2] & [3] & [4] & [6]  & SM & LN & BN & HLN & HBN \\
\midrule
NMI    & 89.46 & 88.65 & 89.48 & 88.70 & \textbf{90.60} & 88.66 & 90.11 & 90.45 & 90.39 & \textbf{90.61}\\
R@1    & 66.67 & 62.46 & 67.02 & 64.52 & \textbf{73.70} & 63.94 & 69.51 & 71.19 & 70.36 & 72.04\\
R@10   & 82.39 & 80.81 & 83.65 & 82.53 & N/A & 80.07 & 84.69 & 85.89 & 85.41 & \textbf{86.25}\\
R@100  & 91.85 & 91.93 & 93.23 & 92.35 & N/A & 90.28 & 92.97 & 93.75 & 93.70 & \textbf{93.80}\\
\bottomrule
\end{tabular}
\end{sc}
\end{small}
\end{center}
\vskip -0.1in
\end{table}

\begin{table}[htb]
\caption{Recall(\%) for In-shop Clothes Retrieval Dataset}
\label{fashion}
\begin{center}
\begin{small}
\begin{sc}
\begin{tabular}{l|cc|ccccc}
\toprule
 & [7] & [8] & SM & LN & BN & HLN & HBN \\
\midrule
R@1    & 53.0 & 62.1 & 78.6 & 79.6 & 80.7 & 80.5 & \textbf{81.1}\\
R@10   & 73.0 & 84.9 & 93.7 & 94.2 & \textbf{94.4} & 94.2 & 94.2\\
R@20   & 76.0 & 91.2 & 95.4 & 96.0 & \textbf{96.1} & \textbf{96.1} & 95.9\\
R@30   & 77.0 & 92.3 & 96.3 & 96.8 & \textbf{96.9} & 96.7 & \textbf{96.9}\\
\bottomrule
\end{tabular}
\end{sc}
\end{small}
\end{center}
\vskip -0.1in
\end{table}

\begin{table}[htb]
\caption{R@1(\%) for Car196 with Different $\alpha$ Values and Embedding Sizes}
\label{embedding}
\begin{center}
\begin{small}
\begin{sc}
\begin{tabular}{c|c|ccccc|c}
\toprule
\backslashbox{\#DIM}{Model} & SM & $\alpha$ = 4 & $\alpha$ = 8 & $\alpha$ = 16 & $\alpha$ = 32 & $\alpha$ = 64 & HBN ($16 \rightarrow 4)$ \\
\midrule
64    & 60.8 & 67.4 & 68.7 & 71.1 & 69.5 & 62.5 & \textbf{74.0}\\
128   & 65.2 & 71.6 & 71.0 & 74.2 & 73.0 & 66.6 & \textbf{77.5}\\
256   & 67.3 & 72.2 & 69.7 & 78.0 & 75.2 & 70.1 & \textbf{80.1}\\
\bottomrule
\end{tabular}
\end{sc}
\end{small}
\end{center}
\vskip -0.1in
\end{table}

\subsection{Evaluation}
Following other existing works on metric learning, we evaluate the clustering quality and the retrieval performance on the images of the test set. 
Following~\citet{song_deep_2016}, all the features are $\ell_2$-normalized before calculating the evaluation metric. 
The normalized features performs slightly better than the unnormalized feature. 

For clustering, the K-Means algorithm is run on all the embeddings of the test samples. 
The number of cluster is chosen to be the number of categories in the test set.
Each test sample will be assigned a cluster index according to which cluster it belongs to. 
Normalized Mutual Information~(NMI)~\citep{schutze2008introduction} between the clustering index and the ground-truth label is used as the metric for clustering. Note that NMI is invariant to the label permutation. 

For retrieval, the performance is evaluated by Recall@K, which is also a widely used metric for this problem. Given a query sample from the test set, K samples from the rest of the test set (or the gallery set for the fashion dataset) with the smallest distance are retrieved. If any retrieved samples is from the same category as the query sample, the recall for this sample is set to 1, otherwise, 0. The reported Recall@K is the average recall on the whole test set. 

We train a classifier on the training dataset with the softmax function and cross-entropy as baseline (SM). For the baseline classifier, in training, the features and the weights are not normalized and $\alpha$ is set to 1. 
4 different versions of classifiers trained with the proposed methods are used for evaluation:
\vspace{-0.6em}
\begin{itemize}
\setlength\itemsep{-0.3em}
\item \textbf{LN}: softmax with $\ell_2$-normalized embedding, $\ell_2$-normalized weights and $\alpha = 16$.
\item \textbf{BN}: softmax with batch normalized embedding, $\ell_2$-normalized weights and $\alpha = 16$.
\item \textbf{HLN}: Heated-up model using $\alpha = 4$ to fine-tune LN.
\item \textbf{HBN}: Heated-up model using $\alpha = 4$ to fine-tune BN.
\end{itemize}
\vspace{-0.7em}

We also compare the proposed method with multiple state-of-the-art metric learning methods. 
Existing literatures are using different base networks and different evaluation protocols. 
For fair comparison, only the methods using GoogleNetV1 as base network and Euclidean distance as the final evaluation metric are listed: [1] Triplet learning with semi-hard negative mining~\citep{schroff_facenet}, [2] Lifted structured loss ~\citep{oh_song_deep_2016}, [3] Learnable Structured Clustering~\citep{song_deep_2016}; [4] Deep Clustering Learning without spectral learning~\citep{law_deep_2017}; [5] Deep Metric Learning with Smart Mining~\citep{harwood_smart_2017} and [6] ProxyNCA~\citep{movshovitz-attias_no_2017}. For the Fashion dataset, we compared with [7] FashionNet \citep{liu2016deepfashion} and [8] Hard-Aware Deeply Cascaded Embedding \citep{yuan_hard-aware_2016}.

\vspace{-1.1em}
\subsection{Metric Learning}
\label{subsec:metric_learning}
\vspace{-0.5em}
The performances of all the methods on all 4 datasets are listed in Tables~\ref{car_196}, \ref{cub200}, \ref{product} and \ref{fashion} respectively. 
The softmax baseline already shows comparable result with many other triplet loss based methods. 
The embeddings trained with either $\ell_2$-normalization or batch normalization improve the performance of the softmax baseline. Since in test phase, all the features are $\ell_2$ normalized before calculating the metric, the performance gain is not coming from a simple $\ell_2$ normalization to the final features. 
Batch normalization works slightly better than $\ell_2$-normalization. 
The ``heated-up'' models (HLN and HBN) show better performance in almost all the metrics compared to the embedding trained with a fixed temperature.

\vspace{-0.9em}
\subsection{Embedding Size and $\alpha$}
\label{subsec:embedding_size}
\vspace{-0.5em}
We further study how different embedding sizes and $\alpha$ values affect the 
retrieval performance. The size of the embedding is chosen in $[64,128,256]$, and the $\alpha$ value in $[4.0, 8.0, 16.0, 32.0, 64.0]$. 
The R@1 metric on the test set with different embedding sizes and $\alpha$ values is reported in Tab.~\ref{embedding}. 
Performances of the feature learned by softmax function without normalization and feature learned from ``heated-up'' model are also given. 
The ``heated-up'' model outperforms all the other models by a significant margin in all cases. 
Between models trained with fixed $\alpha$ values, the model trained with $\alpha = 16$ outperforms others. 

\vspace{-0.9em}
\section{Discussion}
\vspace{-0.9em}
We have discussed 
how the temperature parameter in the softmax function affects the distribution of the embedding in the second last layer of a deep classification model. 
Training with an intermediate temperature will lead to an intra-category compact and inter-category ``spread-out'' embedding which is beneficial for both clustering and retrieval. 
A "heating-up" method is also proposed to further improve the clustering and retrieval performance of the embedding by fine-tuning with a higher temperature. 
Our classifier based approach achieves good performance in metric learning problems with a simpler and more efficient training process than state-of-the-art methods.

\subsubsection*{Acknowledgments}
This material is based upon work supported by the United States Air Force Research Laboratory (AFRL) and the Defense Advanced Research Projects Agency (DARPA) under Contract No.~FA8750-16-C-0166. Any opinions, findings and conclusions or recommendations expressed in this material are solely the responsibility of the authors and does not necessarily represent the official views of AFRL, DARPA, or the U.S. Government.

\bibliography{iclr2019_conference}
\bibliographystyle{iclr2019_conference}

\end{document}